%% file: root.tex
\documentclass[conference]{IEEEtran}
\IEEEoverridecommandlockouts
\usepackage{cite}
\usepackage{amsmath,amssymb,amsfonts}
\usepackage{algorithm, algorithmicx, algpseudocode}
\usepackage{graphicx}
\usepackage{textcomp}
\usepackage{tabularx}
\usepackage{booktabs}
\usepackage{multirow}
\usepackage{url}
\usepackage{hyperref}
\usepackage[capitalise]{cleveref}
\usepackage{subcaption}
\usepackage[dvipsnames]{xcolor}
\DeclareMathOperator*{\argmin}{arg\,min}

\def\BibTeX{{\rm B\kern-.05em{\sc i\kern-.025em b}\kern-.08em
    T\kern-.1667em\lower.7ex\hbox{E}\kern-.125emX}}
\begin{document}

\title{Efficient Data Representation for Motion Forecasting: A Scene-Specific Trajectory Set Approach\\
}

\author{

\IEEEauthorblockN{1\textsuperscript{st}Abhishek Vivekanandan}
\IEEEauthorblockA{%
% \textit{ISPE} \\
\textit{FZI Research Center for Information Technology}\\
Karlsruhe, Germany \\
vivekanandan@fzi.de}
\and
\IEEEauthorblockN{2\textsuperscript{nd} J. Marius Zöllner}
\IEEEauthorblockA{
\textit{FZI Research Center for Information Technology} \\ \textit{KIT - Karlsruhe Institute of Technology}\\
Karlsruhe, Germany \\
zoellner@fzi.de
}
}

\maketitle

\begin{abstract}
% Representing diverse and plausible future trajectories of actors is crucial for motion forecasting in autonomous driving. However, efficiently capturing the true trajectory distribution with a compact set is challenging. In this work, we propose a novel approach for generating scene-specific trajectory sets that better represent the diversity and admissibility of future actor behavior. Our method constructs multiple trajectory sets tailored to different scene contexts, such as intersections and non-intersections, by leveraging map information and actor dynamics. We introduce a deterministic goal sampling algorithm that identifies relevant map regions and generates trajectories conditioned on the scene layout. Furthermore, we empirically investigate various sampling strategies and set sizes to optimize the trade-off between coverage and diversity. Experiments on the Argoverse 2 dataset demonstrate that our scene-specific sets achieve higher plausibility while maintaining diversity compared to traditional single-set approaches. The proposed Recursive In-Distribution Subsampling (RIDS) method effectively condenses the representation space and outperforms metric-driven sampling in terms of trajectory admissibility. Our work highlights the benefits of scene-aware trajectory set generation for capturing the complex and heterogeneous nature of actor behavior in real-world driving scenarios.
Representing diverse and plausible future trajectories is critical for motion forecasting in autonomous driving. However, efficiently capturing these trajectories in a compact set remains challenging. This study introduces a novel approach for generating scene-specific trajectory sets tailored to different contexts, such as intersections and straight roads, by leveraging map information and actor dynamics. A deterministic goal sampling algorithm identifies relevant map regions, while our Recursive In-Distribution Subsampling (RIDS) method enhances trajectory plausibility by condensing redundant representations. Experiments on the Argoverse 2 dataset demonstrate that our method achieves up to a 10\% improvement in Driving Area Compliance (DAC) compared to baseline methods while maintaining competitive displacement errors. Our work highlights the benefits of mining such scene-aware trajectory sets and how they could capture the complex and heterogeneous nature of actor behavior in real-world driving scenarios.
\end{abstract}


% Remove the following section once the paper writing is complete and ensure that the following points are covered.
% \section{Overarching goal of the paper}
% Main key points to cover while writing the parts.
% \begin{itemize}
%     \item \textbf{Core idea: what is the best way to create the sets? map-driven coverage.}
%     \item how many numbers of sets to have?
%     \item How to classify the scenarios based on maps?
%     \item Once the sets are gathered, how to say whether this set is good? 
%     \item What is the baseline to have for such a set? Compare it with a universal small set which has like 10K trajectories which could roughly approximate the driving decisions of the focal agent. Metrics: ADE, FDE.
%     \item Can I use the complete 250K trajectories and prune them? If so, what is the computational cost to perform pruning. 
%     \item In the end, there should be a set where the number of trajectories should remain intact and also should have a representation capability to represent the underlying ground truth.
% \end{itemize}

\input{sections/prologue}
\input{sections/introduction}
\input{sections/related_works}
\input{sections/problem_formulation}
\input{sections/algo}
\input{sections/experiments}
\input{sections/resultsAndDiscussions}
\input{sections/conclusion.tex}

\section*{Acknowledgment}
The research leading to these results was funded by the German Federal Ministry for Economic Affairs and Climate Action and was partially conducted in the project “KI Wissen”. We would also like to thank Max Zipfl and Ahmed Abouelazm for their valuable reviews. Responsibility for the information and views set out in this publication lies entirely with the authors.

\bibliography{references}
\bibliographystyle{IEEEtran}

\end{document}

%% file: sections/prologue.tex
\section{Prologue}
The need for scene-specific trajectory sets arises from the inherent variability in driving environments. Different scenes—such as intersections, roundabouts, or straight roads—impose unique constraints on vehicle movement due to factors like road geometry, traffic rules, and potential interactions with other vehicles. A single set of trajectories cannot capture this diversity effectively because it lacks the contextual information needed to represent plausible behaviors across all scenarios.
For example, at intersections, vehicles may turn left, right, or go straight, each requiring different trajectory predictions based on lane geometry and traffic signals. In contrast, on straight roads, fewer trajectory variations are needed because vehicles typically follow a more predictable path. By generating scene-specific trajectory sets, we can ensure that the predicted trajectories are not only diverse but also plausible within the specific context of each scene.
Moreover, using a single set risks either under-representing possible outcomes (leading to unsafe predictions) or over-representing them (resulting in computational inefficiency). Scene-specific sets strike a balance by focusing computational resources on the most relevant trajectories for each scenario. This approach enhances both safety and performance in autonomous driving systems by ensuring that predictions align more closely with real-world constraints and behaviors.
% \begin{figure}[!t]
%     \begin{minipage}[b]{1.0\columnwidth}
%         \centering
%         \includegraphics[width=0.9\columnwidth]{Figure/output.png}
%         \caption{Trajectory Set: A stateful representation of future actions for target actors derived from training data } % TODO: rename this ugly description
%         \label{fig:my_cure_pic}
%     \end{minipage}

% \end{figure}

% Begin img here +++++++++++++++++++++++++++++
\begin{figure}[!t]
\centering
    \begin{subfigure}[b]{0.8\linewidth}
        \centering
        \includegraphics[width=0.9\linewidth]{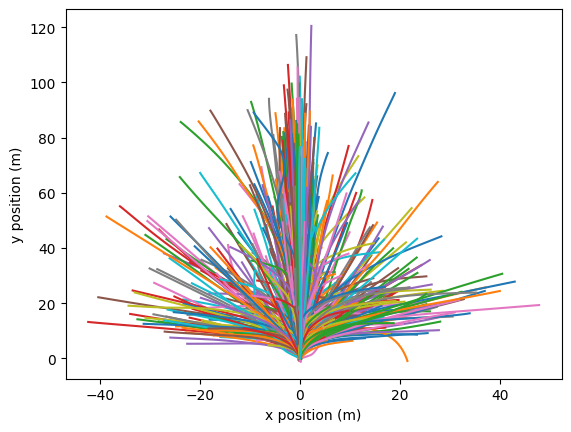}
        \caption{}
        \label{fig:intersection_cluster}
    \end{subfigure}
    \hfill
    \begin{subfigure}[b]{0.8\linewidth}
        \centering
        \includegraphics[width=0.9\linewidth]{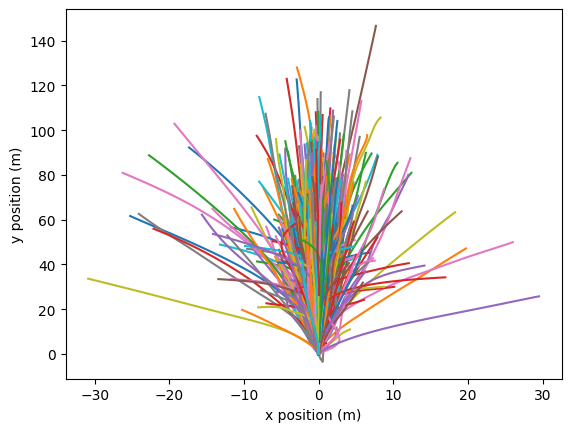}
        \caption{}
        \label{fig:non-intersection_cluster}
    \end{subfigure}
    \caption{ A Set for (a) Intersection Scenes and (b) Non-Intersection Scenes }
\end{figure}
% end img here +++++++++++++++++++++++++++++

%% file: sections/introduction.tex
\section{Introduction}
% 2 problems:
% how traffic participants move in the scene?
In Autonomous Driving, motion forecasting of actors surrounding the ego vehicle has been a persistent challenge due to the evolving complexities associated with a scene, exacerbated by the inherent nature of dynamic interactions between different traffic participants. Capturing such interactions through discrete intent has been successful to some extent, but such representations lack multi-modality to capture the true posterior distribution of the actors. Consequently, motion forecasting demands multi-modality of plausible outcomes, defining the intended actions required to reach a particular goal position \cite{zhao_tnt_2021}.
The associated model uncertainty can result in system behavior that is non-conformant with both physics and scene geometry. In certain situations, this non-conformance can lead to fatal consequences. \cite{noauthor_unacceptably_2022}

Trajectory sets play a crucial role in encoding physics into the motion forecasting layer of an autonomous driving (AD) stack, as demonstrated in the works of \cite{phan-minh_covernet_2020, biktairov_prank_2020, vivekanandan_ki-pmf_2023}. These sets provide a structured representation of the possible future trajectories of traffic participants sampled from the training distribution. By incorporating such trajectory sets into the motion forecasting process, the AD system can generate predictions that are more consistent with the vehicle kinematics and geometric constraints of the scene.

Existing methods compute sets independent of the underlying scene representation (acquired from HD maps) and solely rely on minimizing a cost function based on some metrics. This leads to pertinent gaps about constructing an effective set, whereby the cost function includes a geometric correlation between reachability and the implied metrics.
%However, the effectiveness of trajectory sets in capturing the multi-modality and uncertainty of motion forecasting remains a significant challenge. 
Sets, therefore, must be carefully designed to balance the diversity of plausible outcomes; failure to achieve this balance can lead to suboptimal system behavior arising out of sparsity problems within the output representation space \cite{vivekanandan_ki-pmf_2023}. This is because constructing such an optimal set is a max-min k-dispersion problem, which belongs to the NP-hardness complexity \cite{siciliano_toward_2010}. Through this work, our contributions are twofolds;
% what am i contributing through this work?
\begin{itemize}
\item \textbf{Scene-Specific Trajectory Set Generation:} We introduce a novel algorithm for creating scene-specific trajectory sets by leveraging HD map information and actor state. Our method constructs a directed graph from the vector map, performs a depth-first search to identify goal lanes, and categorizes scenes into intersection and non-intersection bins. This approach enables the generation of tailored trajectory sets that better represent the geometric constraints and likely actor behaviors for each scene type, improving both diversity and plausibility of trajectories.
\item \textbf{Empirical Analysis of Sampling Methods and Set Sizes:} Through extensive experiments on the Argoverse 2 dataset, we investigate the impact of different sampling strategies and set sizes on the diversity and plausibility of the generated trajectory sets. We conduct a comprehensive empirical study comparing metric-driven, random sampling, and our proposed Recursive In-Distribution Subsampling (RIDS) methods across various set sizes (1k to 5k trajectories). The results demonstrate that sampling from a uniform distribution using RIDS outperforms the metric-driven method in terms of retaining more plausible trajectories while maintaining comparable displacement errors. We empirically show that RIDS can effectively condense the full representation space to a manageable set size while preserving coverage and plausibility. Furthermore, our analysis reveals the optimal balance between set size and performance metrics, providing insights for efficient trajectory set generation in downstream motion forecasting tasks.
%\item \textbf{Comparative Analysis of Sampling Methods:} We conduct a comprehensive empirical study comparing metric-driven, random sampling, and our proposed Recursive In-Distribution Subsampling (RIDS) methods across various set sizes (1k to 5k trajectories). Our analysis, using metrics such as minADE and Driving Area Compliance (DAC), demonstrates that RIDS outperforms metric-driven approaches in maintaining trajectory plausibility while providing comparable diversity. We show that random sampling with a 0.2m threshold, followed by RIDS, offers an optimal balance between coverage and admissibility, particularly for non-intersection scenarios where it retains up to 86\% of trajectories in a 5k set.
\end{itemize}

% The implementation of these algorithms involves a step-by-step process:
% Identify the distinct ODDs within the application domain.
% For each ODD, create a dedicated trajectory set that captures the unique characteristics and constraints of the scene.
% Apply the proposed pruning algorithms to each scene-specific set, reducing the set size while preserving the essential information from the underlying distribution.
% Evaluate the performance of the pruned sets in terms of diversity, plausibility, and computational efficiency.
% Integrate the optimized sets into the application life cycle, ensuring seamless transitions between different ODDs.

%% file: sections/related_works.tex
\section{Related Works}
The trajectory prediction problem has been predominantly addressed using \textit{regression} approaches, which aim to predict a set of waypoints that closely resemble the ground truth trajectory. During training, the loss between the ground truth and the model prediction is optimized. %To capture the uncertainty associated with multi-modal trajectories, some methods employ the k-nearest trajectories to compute the loss.
However, \textit{regression} methods often rely on anchors to enforce diversity between trajectories and require anchors for preventing the models against mode collapse \cite{cui_multimodal_2019, varadarajan_multipath_2021}.  Regression-based approaches without explicit integration of prior knowledge leads to predictions which go off-road \cite{bahari_vehicle_2022} without respecting the neither the underlying kinematics nor the environment \cite{casas_importance_2020, vivekanandan_ki-pmf_2023}.%may not scale well in complex urban scenarios due to the explosion of anchor numbers, failing to capture multimodality thereby reducing prediction quality. 

\textbf{Trajectory Sets in Planning and Prediction}
Trajectory sets have emerged as a choice in motion prediction, particularly in autonomous driving, as they model the uncertain future as a distribution over a compact set of possible trajectories for each agent. Predominantly, trajectory sets or trim trajectories have been used in planning \cite{branicky_path_2008} to resolve for a valid path at a constant time and memory. This flexibility has demonstrated state-of-the-art performance \cite{phan-minh_covernet_2020, schmidt_reset_2023, biktairov_prank_2020} by reformulating the problem as a \textit{classification} where each of the trajectories in the set represents an individual anchor \cite{varadarajan_multipath_2021}. This effectively avoids mode collapse, which can occur with models when trained with a regression loss. While traditional methods like Bayesian approaches \cite{catanach_computational_nodate} can be computationally intensive and may pose limitations for real-world applications, whereas trajectory sets offer a practical and empirically superior alternative.

%The development of trajectory sets and associated pruning methods can be traced back to the DARPA challenge and the work of \cite{branicky_path_2008}. Pruning algorithms were established as necessary to maintain constant compute-to-scale time and compensate for the limited computational resources available at that time. 

%\textbf{Compute to Scaling and Complex Urban Scenarios}
%The concept of “compute to scaling” is crucial when considering the deployment of models in production environments, as it refers to the model's ability to scale efficiently in relation to the number of agents, especially in complex urban scenarios. Although it is relatively straightforward to develop a model that can maximize a distribution, ensuring that the model can scale with the number of agents in complex situations remains a significant challenge.
%In simpler driving situations, such as on highways, the computational demands are lower due to fewer vehicle choices compared to more complex driving scenarios. This underscores the importance of trajectory sets in motion forecasting, as they strike a balance between computational efficiency and the ability to scale in complex environments.

\textbf{Multiple Sets for Improved Representation}
Single-set representations often fail to capture the true posterior distribution of ground truth probabilities in real-world scenarios due to sparsity and model expressivity issues \cite{vivekanandan_ki-pmf_2023, zhang_understanding_2017}. We propose using multiple subsets to create a denser representation space that better approximates ground truth probabilities and trajectory existence. The effectiveness of these sets depends on scene clustering methods, which can be supervised or semi-supervised \cite{zipfl_self_2022}. Various approaches exist, from using traffic actors' state vectors with criticality metrics \cite{zipfl_fingerprint_2022, schutt_clustering-based_2023, westhofen_criticality_2023} to clustering based on agent trajectories \cite{bernhard_optimizing_2021}. Our method employs an L-norm metric between lane centerlines and the target actor's current state, avoiding the look-ahead bias present in other methods that require full actor state representation to create cluster bins. 

% Although IE is expected to perform well in maximizing the probability of being close to the ground truth, real-world data and minADE measurements between training and validation data indicate otherwise. This discrepancy led to the development of experiments to understand how trajectories fare with randomly sampled sets and pruning based on different scenarios.

%\textbf{Map Information Integration and Dataset-Agnostic Approach}
In our previous work \cite{vivekanandan_ki-pmf_2023}, we investigated the approach of pruning trajectory sets based on their intersection with non-drivable regions. However, the generated set lacked priors knowledge with map information, leading to a significant number of trajectories being discarded due to non-plausible states. To address this limitation and build upon the new concept of multiple sets, we propose an alternative dataset-agnostic methodology that understands map information and produces multiple sets, helping to overcome sparsity issues and maintain an optimal balance between diversity and admissibility, as discussed by Park et al. \cite{park_diverse_2020}.

%% file: sections/problem_formulation.tex
\section{Problem Formulation}
We begin by establishing the formal notations that will be consistently utilized throughout this document. Consider a full set of trajectories captured from the training distribution as $T_{full}$. Our objective is to deduce a subset $T' \subset T_{full}$ that maximizes the diversity and plausibility of trajectories for a cluster of scenes $M_{i}$.

\begin{equation}
M = \{M_i \mid i \in \mathbb{Z}^+, 1 \leq i \leq n\}
\end{equation}

Let $M$ be the set of all scene clusters where $M_i$ represents an individual cluster of scenes and $i$ is a positive integer (denoted by $\mathbb{Z}^+$) with $n$ being the total number of clusters.
%Clustering individual scenes into one of the clusters is done through the Alg. \ref{alg1}.
$|T'|$ denotes the cardinality of a set $T'$.
A trajectory $T_i$ of an actor $a_i$ is represented by a sequence of states, as shown in \cref{eq:combined_traj}, where $s_i^t$ represents the center point location $(x, y)$ of $a_i$ at a time $t$ in a Cartesian coordinate frame.

\begin{equation}
    T_i = \left \{ \underbrace{s_i^{-t_{obs}}, s_i^{-t_{obs} + 1}, s_i^{-t_{obs} + 2}, ...s_i^{0}}_{\text{Past states}}, \underbrace{ s_{i}^{1}, s_{i}^{2}, ..., s_{i}^{t_{fut}}}_{\text{Future states}}\right \}
    \label{eq:combined_traj}
\end{equation}

Here, $t_{\text{obs}}$ represents the end of the observation window, and $t_{fut}$ represents the end of the prediction horizon. This notation aligns with the formalism used throughout the paper, particularly focusing on the representation of trajectories as sequences of states over time.

While promoting diversity ($D$) among the trajectories within  $T'$, it is important to note that coverage alone is insufficient to ensure accurate and plausible output. $T'$ should also be realistic and plausible ($P$), and therefore should not contain states that are physically infeasible to reach for a given scene $S$ in $M_{i}$. 
%A scene, represented as an HD map, encompasses both past $S_{\text{past}}$ and future $S_{\text{future}}$ observations of all traffic participants. 
Additionally, $S$ includes lane information that delineates the probable motion paths an actor can choose to reach a goal position. Within this work, a Focal Track ($FT$) is of an object type car. 
%The future predicted trajectory of a target actor $a_i$ is represented as a sequence of predicted center point coordinates. This sequence is denoted by $S'_i$ for the actor $i$. The set of all actors' future trajectories is denoted by $S'$. The notation for the future trajectory of the actor $i$ is given by:

\begin{equation}
\max_{T' \subset T_{full}} \left[ D(T') + P(T'|M_i) \right]
\label{eqn:maximization_of_set}
\end{equation}

Thus, we aim to maximize \cref{eqn:maximization_of_set} where we search for a set which not only optimizes for the coverage metric but also be diverse enough to sustain the environmental constraints without getting pruned \cite{vivekanandan_ki-pmf_2023}; a balance between diversity and plausibility. 

%% file: sections/algo.tex
\section{Scene-Specific Set Generation}

% Begin algo here +++++++++++++++++++++++++++++
\begin{algorithm}
\caption{Scene-Specific Set Generation}
\label{Alg:alg1}
\begin{algorithmic}[1] % The [1] option enables line numbering
\Require Vector-Map ($m$), State ($s_{i}$) Information of Focal Track $FT$
\For{each Scene}
    \State Construct a directed graph $G$ from $m$
    \State {$s_{i}$}, $\theta$ $\leftarrow$ Query for the focal track states, where $\theta$ is the heading angle
    \State $R \leftarrow$ Calculate Rotation Matrix with $\theta$
    \State $N_{nearest} \leftarrow$ Filter for edges with parallel direction vectors to $FT$
    \State $n_{source} \leftarrow$ Find the source node using L1-norm between {$s_{i}^{0}$} and $N_{nearest}$
    \State $L_{goal} \leftarrow$ Perform DFS on all the nodes in $N_{nearest}$
    \For{$l$ in $L_{goal}$}
        \If{$l$ is in an Intersection}
            \State \Return Trajectory, $T_{scene} = 1$
        \Else
            \State \Return Trajectory, $T_{scene} = 0$
        \EndIf
    \EndFor
    \State \Return None
\EndFor

\end{algorithmic}
\end{algorithm}

% end algo here +++++++++++++++++++++++++++++

\subsection{Algorithm}
The deterministic goal sampler operates on the scene \(S\) and the heading angle \(\theta\) w.r.t the Focal Track to generate a set of \(K\) \textcolor{violet}{goal lanes}, denoted as \(L_{goal}\).
This operation can be succinctly represented by the following \cref{eqn:defining_goal}:
\begin{equation}
L_{goal} = f(S, \theta)
\label{eqn:defining_goal}
\end{equation}

Given a vector map of $S$ and the state information of $FT$, the algorithm constructs a directed graph $G$ from the map. The last observation state ($s_i^{0}$) of the $FT$ is used to find the nearest nodes by filtering for edges with parallel direction vectors to that of the $FT$. This helps us find for the edges which have a higher affinity to the $FT$'s travelling direction, mathematically given through the dot product between the two normalized vectors (thresholding it at an angle of 10$^{\circ}$). Within this pool of filtered edges, we choose a \textcolor{violet}{source node} which fulfills the property of being closer to $FT$; found by calculating the L1-norm between $s_i^{0}$ and the nearest nodes. 

A Depth-First Search (DFS) is performed on each of the nearest nodes to find the \textcolor{violet}{goal lanes}. For each goal lane, the algorithm checks if it is in an intersection. If any one of the lanes $l$ in $L_{goal}$ is present in an intersection, the trajectory is returned with a scenario tag of 1; otherwise 0. This process allows for the categorization of scenes into two bins based on the presence of goal lanes in an intersection.
Running \cref{Alg:alg1} across the training dataset yields two clustered sets that reflect the local representation of the scene, as shown in \cref{fig:intersection_cluster} and \cref{fig:non-intersection_cluster}. The trajectories corresponding to the scenes are normalized to $FT$'s last observation state using the rotation matrix $R$. 

As we will see in the upcoming chapters, how this clustering maximizes the probability of finding trajectories closer to the ground truth without compromising on the shortcomings of earlier approaches which utilize a single set. To optimize the clustered sets further, we need to determine two key parameters: \textit{the optimal set size} and \textit{the sampling strategy}. The sampling approach plays a crucial role in reducing the clustered sets to an optimal size suitable for the intended application.

\section{Metrics}
% insert points about minADE, minFDE, Driving Area compliance

The distance between trajectories within a set is quantified using Euclidean metrics, such as the minimum Average Displacement Error (minADE) and the minimum Final Displacement Error (minFDE). A diverse set of trajectories serves as a representative sample of the underlying population, capturing its intrinsic characteristics. By assessing the properties of each trajectory within the set, we can quantify its individual contribution to the overall diversity. To measure the plausibility of the set with respect to the drivable regions of a scene, we employ the Driving Area Compliance (DAC) score, which measures the compatibility of a set for a given scene. Therefore,
\[
DAC = \frac{|T'|-n}{|T'|}
\]
where, $|T'|$ represents the given set size and $n$ denotes the number of non-compliant trajectories which goes off-road. A higher DAC score provides an indication of the set's density in relation to the scene.  We use the same methodology as described in \cite{vivekanandan_ki-pmf_2023}
to get the DAC value per scene.
By integrating the DAC and Euclidean metrics, we enhance our assessment of the set's quality, which hinges on its closeness to the ground truth trajectories. A higher DAC score signifies that more trajectories adhere to drivable areas, whereas lower minADE values indicate closer alignment with the ground truth. Employing both diversity and scene compliance metrics ensures that the optimized set of trajectories maximizes the representativeness of the set, providing a comprehensive portrayal of the true population dynamics w.r.t the map geometry.

To evaluate the quality of our constructed trajectory set, we establish a theoretical Lower Bound (LB) through comparison with the $T_{full}$ which provides us with the required query trajectories \footnote{We use both query trajectory and ground truth trajectory interchangeably throughout this paper}. The LB computation process operates as follows: for each query trajectory $q \in T_{full}$, we determine the closest matching trajectory $T^*$ for a given $T'$ by calculating minADE, formally expressed as:
% Begin img here +++++++++++++++++++++++++++++
\begin{figure*}[ht] 
    \centering
    % First row of images
    \begin{subfigure}[b]{0.325\textwidth}
        \centering 
        \includegraphics[width=0.9\linewidth]{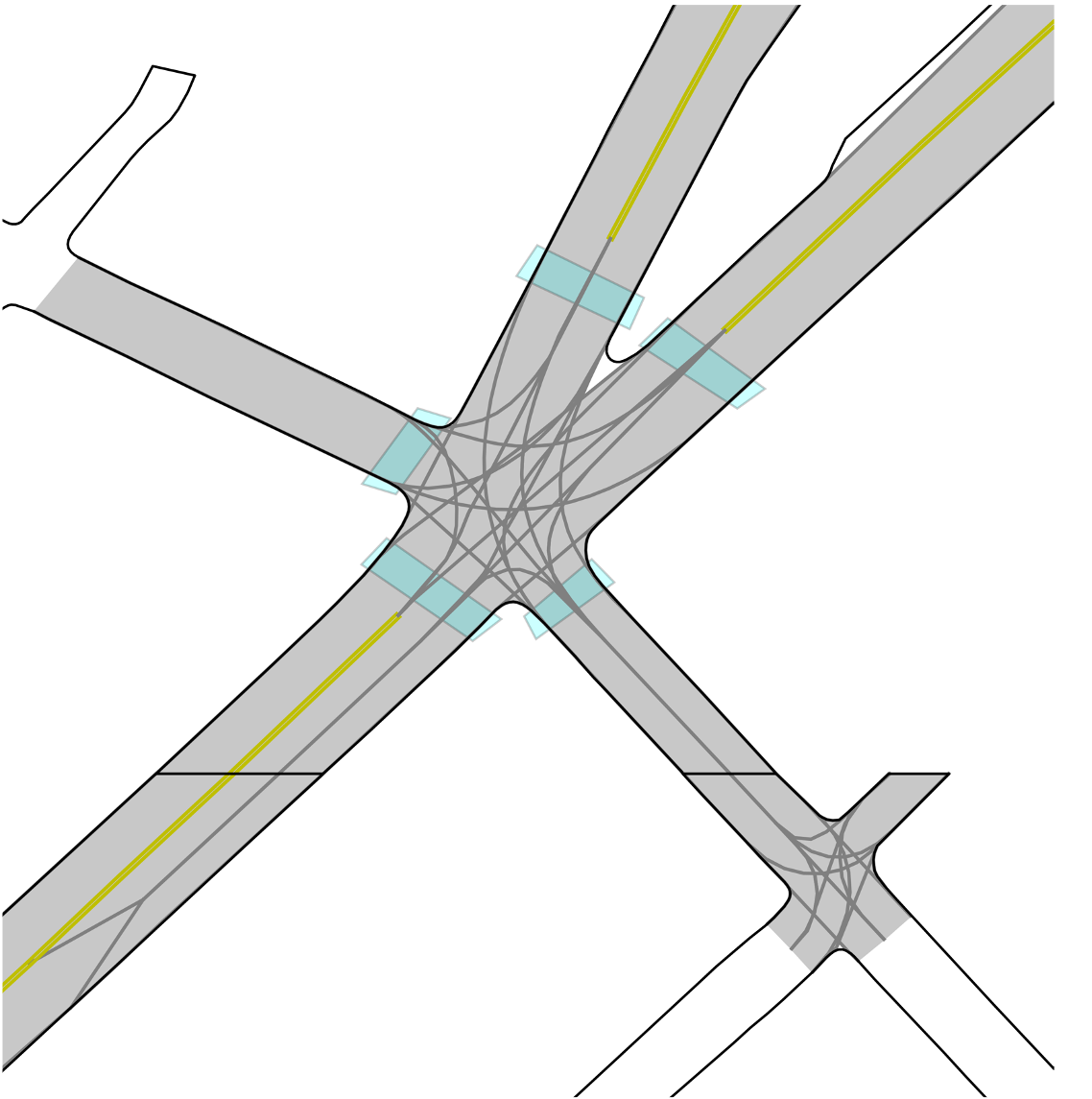}
        \caption{Drivable Regions}
    \end{subfigure}
    % \hspace{2mm} 
    \begin{subfigure}[b]{0.325\textwidth}
        \centering
        \includegraphics[width=0.9\linewidth]{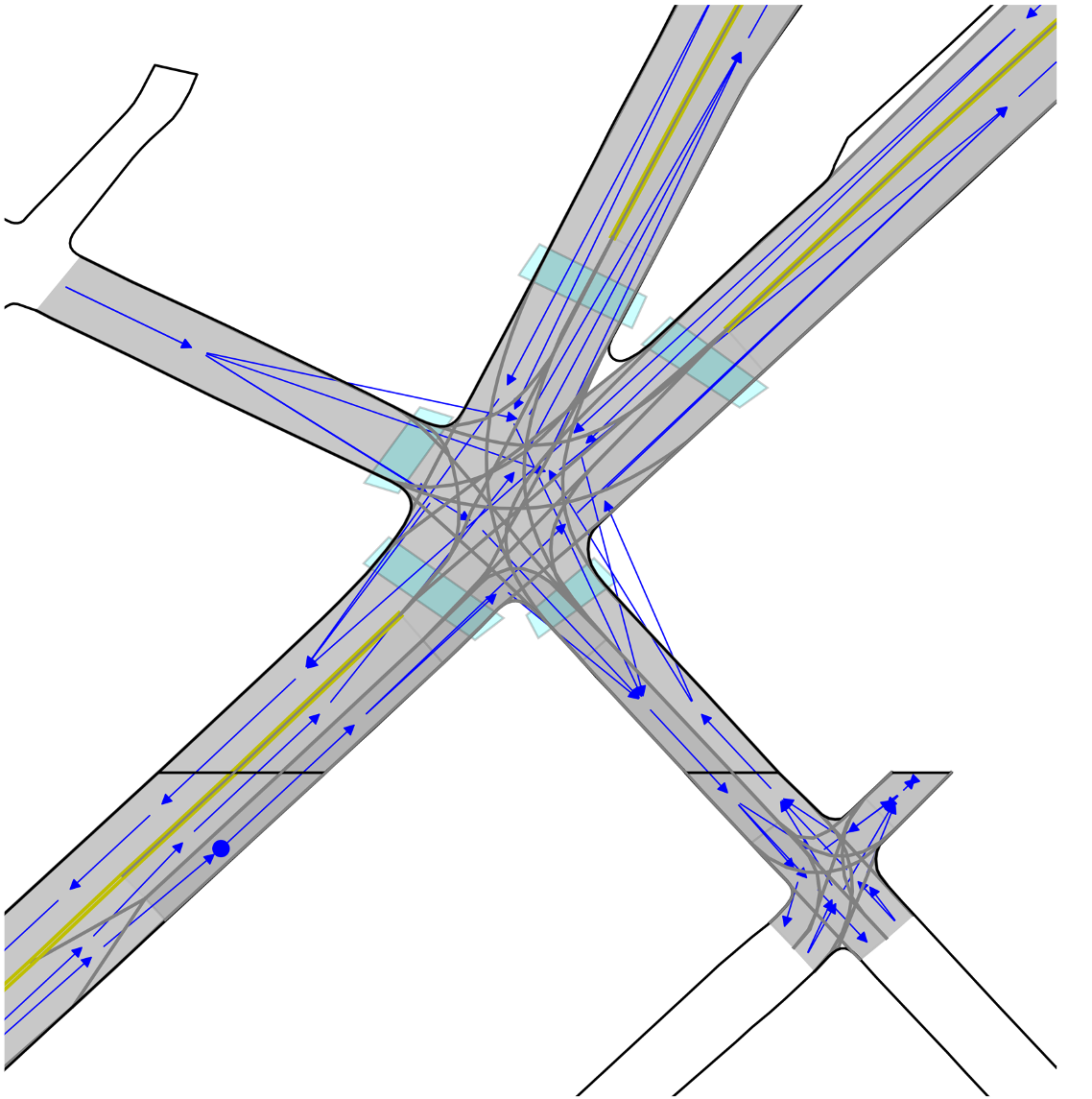}
        \caption{Node Construction for Traversable Lanes}
    \end{subfigure}
    % \hspace{2mm}
    \begin{subfigure}[b]{0.325\textwidth}
        \centering
        \includegraphics[width=0.90\linewidth]{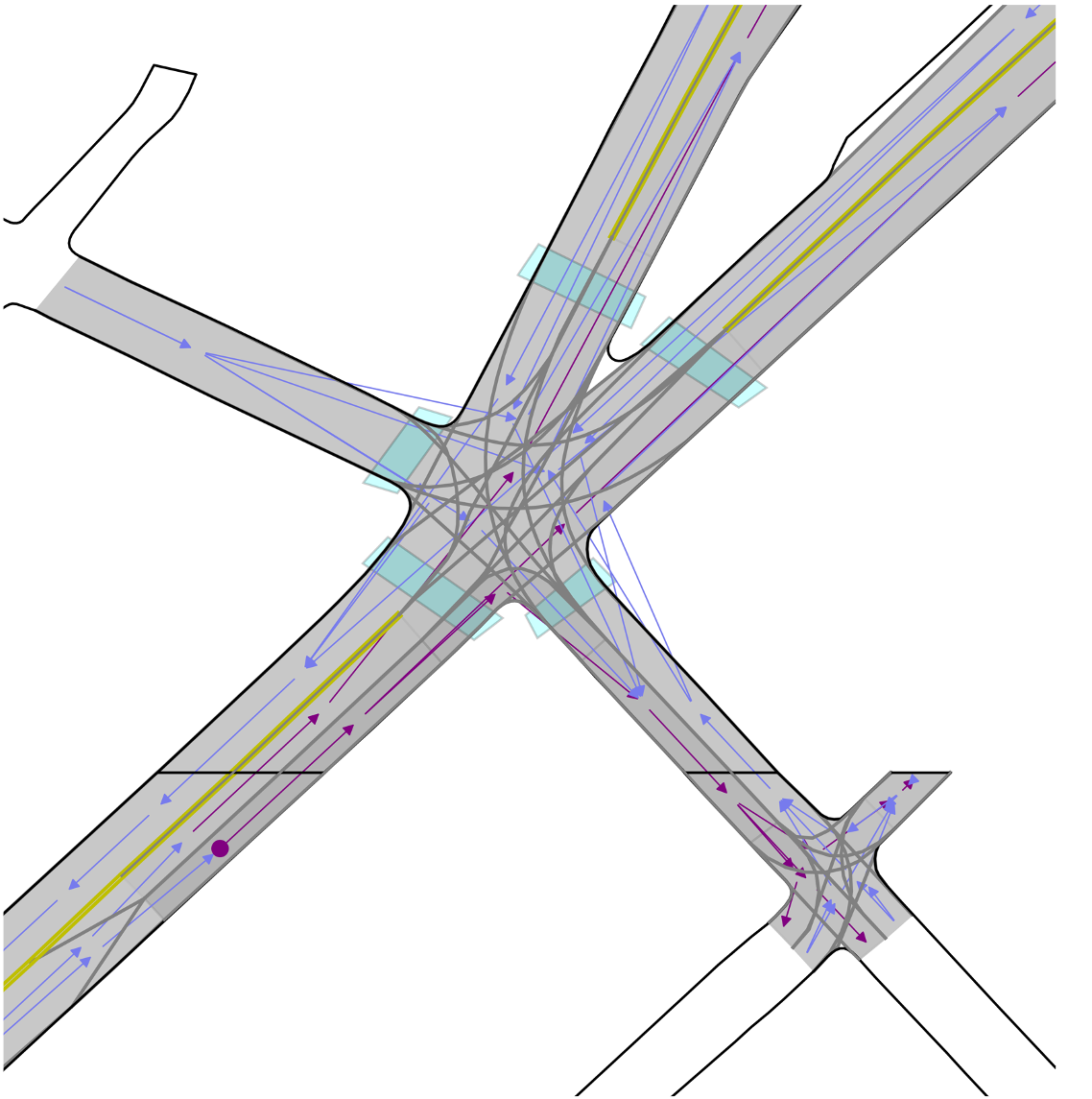} 
        \caption{Reachable Lanes}
    \end{subfigure}
    
    \begin{subfigure}[b]{0.5\textwidth}
        \centering
        \includegraphics[width=0.80\linewidth]{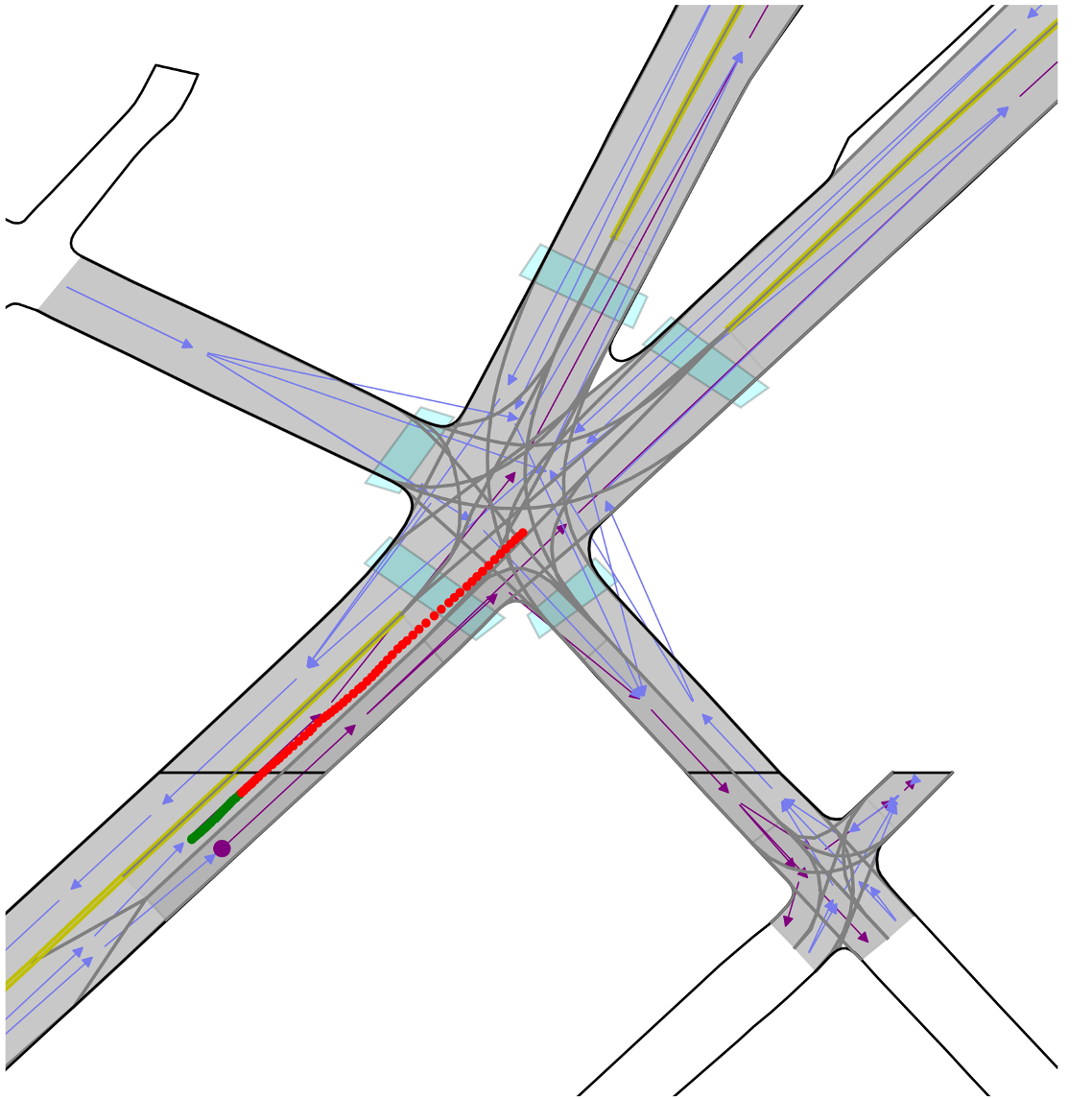} 
        \caption{Complete Scene with Actor Tracks}
    \end{subfigure}
\caption{Steps involved in clustering the scenes based on the road geometry and actor states are shown here. \textcolor{green}{Past trajectory} and \textcolor{red}{Future trajectory} of an FT are illustrated. In (a), \textcolor{gray}{Grey connections} denote the connectivity between various lanes at an intersection. In (b), \textcolor{blue}{Blue edges} depict the transformation of lane connections into node edges. Using \cref{Alg:alg1}, the \textcolor{violet}{Source node}, which corresponds to the closest node relative to the last observed state of the FT, is identified. Finally, the edges in (c) represent the plausible lanes the FT could potentially select.}
\end{figure*}
% end img here +++++++++++++++++++++++++++++
%Begin Equation ++++++++++++++++++++++++++++++++++
\begin{equation}
T^* = \argmin_{q \in T_{full}} \text{ADE}(q, T')
\end{equation}
The LB is then computed as the mean of these minimum displacement errors across all query trajectories:
\begin{equation}
\text{LB.minADE} = \frac{1}{|T_{full}|} \sum_{q \in T_{full}} \min_{T \in  T'} \text{ADE}(q, T)
\end{equation}
%End Equation ++++++++++++++++++++++++++++++++++

This metric provides a quantitative measure of how well our constructed set captures the diversity and representativeness of the full trajectory distribution; with lower values indicating better coverage of the possible motion patterns.
The \textit{LB. minADE w/ $T_{full}$} is an important measure
for which the $q$ originates from their respective scene types (intersection or non-intersection), thereby representing ground truth (GT) samples from the scene-specific cluster. The constructed set, however, draws trajectories from $T_{full}$, enabling a comprehensive cross-evaluation of the set's performance across different scenarios.
\begin{equation}
Q_s \subset M_{i}, \quad T' \subset T_{full} %, \quad S \subset T_{full}
\end{equation}
where $Q_s$ represents the query trajectories from a specific scene type $s$. %and $S$ denotes our constructed set.
The establishment of this baseline metric is crucial, as it supports our hypothesis that trajectory sets sampled from $T_{full}$ will inherently exhibit higher displacement errors compared to scene-specific sets. This phenomenon can be attributed to the inherent diversity and potential noise in the full distribution, whereas scene-specific sets benefit from more focused and contextually relevant trajectory patterns. 
%Mathematically, we expect:
%\begin{equation}
%\text{LB}(S_{scene}) < \text{LB}(S_{full})
%\end{equation}
%where $S_{scene}$ and $S_{full}$ represent scene-specific and full-distribution sets, respectively.

%% file: sections/experiments.tex
\section{Experiments}
We use the motion forecasting dataset from Argoverse 2 (AV2)\cite{wilson_argoverse_nodate}, which contains 199,489 training samples or scenes and 20,000 validation samples. The construction and analysis of the validation samples are refrained from, as that would lead to look-ahead bias. Therefore, all the analysis is performed on the training distribution unless explicitly stated otherwise.

The dataset contains HD maps as a lane graph, from which we construct a directed Graph similar to VectorNet polyline \cite{gao_vectornet_2020} representation. Every scene has a target actor called as a focal track ($FT$), whose observation window satisfies the requirement of possessing full state information throughout the complete sampling horizon. For AV2, the sampling horizon is 6s for the future, with a 5s observation history; sampled at 10Hz. A Rotation matrix $R$ is calculated using the heading angle and the last XY coordinate at a time step $t_{\text{obs}}$. The trajectory $T_i$ is transformed into a local coordinate system (actor-centric) by applying $R$, which is better suited to represent trajectories.

%A trajectory set for the complete training distribution ($T_{full}$) is constructed by parsing through each scene and querying for the focal track's future position. Since each of the trajectory is normalized with their respective $R$, we get a set which resembles \cref{fig:my_cure_pic}. 

\subsection{Evaluation of sets on $T_{full}$ }

% Begin Table ++++++++++++++++++++++++++++++++++
\begin{table}[!t]
\centering
\caption{Comparison of coverage-based approach vs. random sampling from the training distribution. $(\downarrow)$ indicates lower is better, while the inverse holds true to $(\uparrow)$. The notation \textit{0.2m} with Ran-Samp represents the distance threshold applied when calculating metrics for randomly sampled trajectories}
\resizebox{\columnwidth}{!}{%
\begin{tabular}{@{}llll@{}}
\toprule
\textbf{Method} & \textbf{Set Size} & \textbf{LB. minADE $(\downarrow)$} & \textbf{DAC}$(\uparrow)$ \\ 
\midrule
                        & 1k    &  0.633   & 0.525 \\
                        & 1.5k  &  0.587   & 0.520 \\
                        & 2k    & 0.558   & 0.499 \\
\textbf{Metric Driven}  & 2.5k  & 0.525   & 0.494 \\
                        & 3k    & 0.511 & 0.490 \\
                        & 4k    & 0.482    & 0.481 \\
                        & 5k    & 0.451  & 0.486 \\ 
\midrule
                            & 1k \textit{(854)}     & 0.771 \textit{(0.782)} & 0.611 \textit{(0.673)} \\
                            & 1.5k \textit{(1246)}  & 0.710 \textit{(0.737)} & 0.627 \textit{(0.662)} \\
                            & 2k \textit{(1642)}    & 0.661 \textit{(0.667)} & 0.607 \textit{(0.664)} \\
\textbf{Ran-Samp} (0.2m)    & 2.5k \textit{(2067)}  & 0.617 \textit{(0.621)} & 0.615 \textit{(0.678)} \\
                            & 3k \textit{(2450)}    & 0.598 \textit{(0.603)} & 0.608 \textit{(0.682)} \\
                            & 4k \textit{(3237)}    & 0.559 \textit{(0.564)} & 0.606 \textit{(0.671)} \\
                            & 5k \textit{(4023)}    & 0.532 \textit{(0.538)} & 0.634 \textit{(0.675)} \\ 
\bottomrule
\end{tabular}
}
\label{table:1}
\end{table}
% End Table ++++++++++++++++++++++++++++++++++. 
Validating the effectiveness of the sets is crucial to comprehend the influence of two primary parameters which govern diversity and plausibility. We employ the \textit{metric-driven} method \cite{schmidt_reset_2023} and a random sampling to demonstrate the efficacy of our proposed scene-specific set generation concept. Similar to the Covernet \cite{phan-minh_covernet_2020}; \textit{metric-driven} method also employs a minADE driven bagging algorithm to reduce each of the clustered sets, we have acquired from the \cref{Alg:alg1}. Random sampling (\textit{Ran-Samp}) is used as a baseline, and we report the lowest value as it exhibits a variance of $\pm \text{2\%}$. By comparing the performance of our method against the baseline and analyzing the lower bounds, we gain insights into how effective our set construction is in capturing diverse and plausible trajectories. 

%We compute a theoretical Lower Bound (LB) which measures the quality of the constructed set validated against the trajectories from the full training distribution ($T_{full}$). For a single query trajectory from $T_{full}$, we find the closest trajectory from the set which shows the lowest displacement error between them. For all the query trajectories present in the $T_{full}$ we perform the same action and report the mean value which represent the LB. minADE for the corresponding set.  
%This lower bound is derived from the minADE between the query (ground truth) trajectory and the closest trajectory within the set. In this case, every trajectory present within the set is validated with every other trajectory and the mean value is reported.

\begin{figure}
    \centering
    \includegraphics[width=0.90\columnwidth]{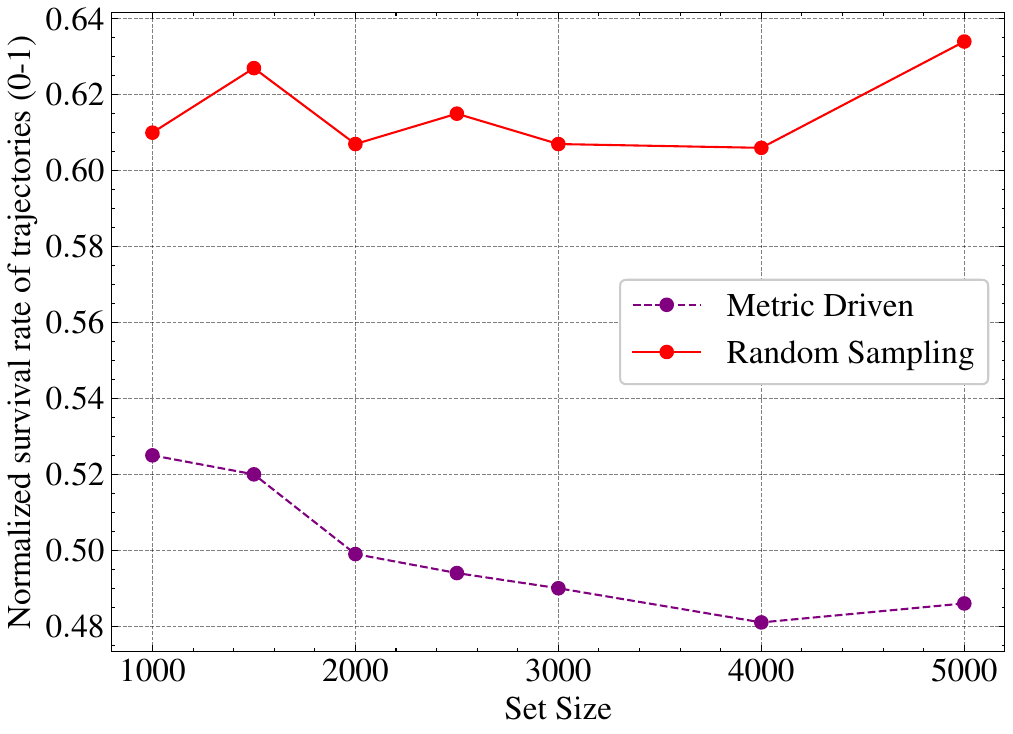}
    \caption{The randomly sampled sets can retain more number of trajectories when compared with a metric-driven set, making the former a reliable way to represent more plausible states}
    \label{fig:comparison-md-rs}
\end{figure}

\cref{table:1} illustrates the effect of \textit{metric-driven} sets when compared with randomly sampled sets generated from the training distribution. As the set size increases, the LB of the metrics linearly decreases, this is attributed to the presence of more trajectories which lie closer to a given $q$, thereby providing good coverage. Sets generated through \textit{the metric-driven} method perform well as expected, since the cost function enforces diversity between the trajectories in the set when compared with \textit{Ran-Samp}. 
On the other hand, from \cref{fig:comparison-md-rs} emerges a wholly different picture. The normalized rate of surviving trajectories, that which are compliant with the scene, is substantially higher for the \textit{Ran-Samp} and amounts to $10\%$ more trajectories irrespective of the set size. \\
The \textit{Ran-Samp} method, however, is not without its drawbacks. It includes trajectories that overlap and are deemed redundant, which could be eliminated without adversely affecting the metrics. To address this, we introduce a thresholding rate of $0.2$m, enabling the removal of redundant trajectories that lie close to each other within the \textit{Ran-Samp} set. This refinement process results in a more lean set, reducing the number of trajectories by approximately $15\%$ compared to the original set size, as can be seen from \cref{table:1}. Moreover, this process improves the DAC rate while preserving an equivalent minADE when juxtaposed with simple \textit{Ran-Samp} set.

\subsection{Evaluation of scene-specific sets }
Utilizing \cref{Alg:alg1}, we generate two distinct clustered sets: one pertaining to intersections and the other to non-intersections. Subsequently, for each clustered set, adhering to their corresponding sampling criteria, we performed extensive analysis, as can be seen from the \cref{table:Comparison_between_different_sets}.

The \textit{LB.minADE} column of \cref{table:Comparison_between_different_sets}, when evaluated against the results from the \cref{table:1} shows similar correlation between the internal representation of trajectories (sets' diversity) and the set size. As we incrementally increase the set size from 1k to 5k, the \textit{LB.minADE} shows a consistent decrease for both methods (\textit{metric-driven} and \textit{Ran-Samp}) and across both scenario types (Non-intersection and intersection set). This suggests that the generation of scene-specific sets increases the probability of finding more trajectories that lie closer to the ground truth.

%In this, we assume that  query trajectory comes from the respective non-intersection/intersection scenes thus representing GT from the scene distribution. Where as the constructed set samples the trajectories from the $T_{full}$ this ensures that we have a cross evaluation about the quality of the sets performance. Establishing a baseline with this metric is essential, as it hypothesizes that a set derived from the full training distribution is expected to perform less effectively compared to the displacement metrics of sets tailored to specific scenes.

%represents a set of potential future states (GT trajectory) for which we are interested in evaluating how close we can find a trajectory from our generated sets (1K, 2K) using specific methods such as metric-driven, random sampling or RIDS. Here, the sampling distribution used to construct our sets comes from the full training distribution rather than being tailored to a specific scene type. This ensures that we establish a baseline for how a set constructed from the full distribution compares with those specific to different scene types (non-intersection/intersection). The intuition behind reporting \textit{LB. minADE w/ set} is to provide a fair comparison between these methods across varying scenes and identify the most suitable approach for each type. Establishing a baseline with this metric is essential, as it hypothesizes that a set derived from the full training distribution is expected to perform less effectively compared to the displacement metrics of sets tailored to specific scenes. 

The \textit{DAC} metric exhibits relative stability across different set sizes for each method and scenario combination. The \textit{Ran-Samp} method consistently shows higher DAC values compared to \textit{metric-driven}, suggesting that it is more effective in generating a set where a larger proportion of the trajectories remain plausible. When comparing the two scenario types, it is evident that both methods yield poorer performance in intersection scenarios compared to non-intersections, as measured between corresponding  \textit{LB.minADE} metrics. This outcome is anticipated due to the increased complexity and diversity of potential trajectories at intersections. However, the performance degradation from non-intersection to intersection scenarios is less evident for \textit{Ran-Samp} than for \textit{metric-driven}. As set sizes increase, the displacement error narrows, and the number of surviving trajectories experiences a significant increase, a trend that is also observable in the intersection sets.

We also created a new subsampling algorithm, \textit{RIDS} or \textit{Recursive In-Distribution Subsampling}, where we remove overlapping trajectories from a set constructed using \textit{Ran-samp}. If two trajectories share proximity to each other and lie within a threshold of \textit{0.2m} to each other, we consider this trajectory redundant and arbitrarily remove one of them. This leaves us with a set which is underweighted; to counter this, we perform subsampling from the training distribution and recursively perform this step of removing and adding trajectories until the desired $T'$ is reached.
% Begin img ++++++++++++++++++++++++++++++++++
% End img ++++++++++++++++++++++++++++++++++ 
% Begin Table ++++++++++++++++++++++++++++++++++
\begin{table*}{}
\centering
\vspace{0.2cm}
\caption{Quantitative comparison of Scene-specific sets with different methods on Argoverse 2 training distribution}
\begin{tabularx}{1.75\columnwidth}{lX|XXX|XXX}
\toprule
& \multicolumn{1}{c}{} & \multicolumn{3}{c}{\textbf{Non-Intersection Set}} & \multicolumn{3}{c}{\textbf{Intersection Set}} \\
\midrule
\textbf{Method} & \textbf{Set Size} & \textbf{LB. minADE} $(\downarrow)$ & \textbf{LB. minADE w/ $T_{full}$} $(\downarrow)$ & \textbf{DAC}$(\uparrow)$ & \textbf{LB. minADE} $(\downarrow)$ & \textbf{LB. minADE w/ $T_{full}$} $(\downarrow)$ & \textbf{DAC}$(\uparrow)$ \\
\midrule
                        & 1k & 0.592 & 0.670 & 0.694 & 0.722 & 0.735 & 0.518 \\
                        & 1.5k & 0.531 & 0.628 & 0.673  & 0.667 & 0.672 & 0.500 \\
                        & 2k & 0.500 & 0.593 & 0.677  & 0.621 & 0.639 & 0.490 \\
\textbf{Metric Driven}  & 2.5k & 0.472 & 0.567 & 0.664 & 0.599 & 0.608 & 0.484 \\
                        & 3k & 0.443 & 0.544 & 0.704 & 0.575 & 0.585 & 0.477 \\
                        & 4k & 0.416 & 0.527 & 0.692 & 0.547 & 0.553 & 0.468 \\
                        & 5k & 0.388 & 0.491 & 0.687 & 0.510 & 0.528 & 0.469 \\
\midrule
                        & 1k & 0.727 & 0.830 & 0.827 &  0.874 & 0.892 & 0.691 \\
                        & 1.5k & 0.673 & 0.748 & 0.831 & 0.789 & 0.813 & 0.664 \\
                        & 2k & 0.601 & 0.685 & 0.823 & 0.729 & 0.746 & 0.663 \\
\textbf{Ran-Samp}       & 2.5k & 0.568 & 0.654 & 0.842 & 0.690 & 0.709 & 0.678 \\
                        & 3k & 0.549 & 0.630 & 0.833 & 0.659 & 0.688 & 0.692 \\
                        & 4k & 0.498 & 0.597 & 0.835 & 0.606 & 0.624 & 0.681 \\
                        & 5k & 0.453 & 0.552 & 0.863 & 0.579 & 0.587 & 0.679 \\
\midrule
                        & 1k & 0.712 & \--- & 0.824 &  0.866 & \--- & 0.689 \\
                        & 1.5k & 0.661 & \--- & 0.827 &  0.779 & \--- & 0.658 \\
                        & 2k & 0.591 & \--- & 0.826 &  0.717 & \--- & 0.656 \\
\textbf{RIDS}           & 2.5k & 0.557 & \--- & 0.836 &  0.676 & \--- & 0.650 \\
                        & 3k & 0.532 & \--- & 0.829 &  0.651 & \--- & 0.682 \\
                        & 4k & 0.488 & \--- & 0.839 &  0.598 & \--- & 0.688 \\
                        & 5k & 0.455 & \--- & 0.850 &  0.572 & \--- & 0.681 \\
\bottomrule

\end{tabularx}
\label{table:Comparison_between_different_sets}
\end{table*}
% End Table ++++++++++++++++++++++++++++++++++

%% file: sections/resultsAndDiscussions.tex
\section{Results and Discussions}
The ultimate goal of this work is to find the answer to the following question: \textit{What is the best conceptual framework that can maximize the probability of finding trajectories closer to the ground truth?} This led us to investigate and propose a scene-specific set generation method, which, when viewed through the lenses of displacement errors and DAC, gives better performance compared to an uninformed set. The next step is to find the sampling method; From \cref{fig:comparison-md-rs}, random sampling provides preliminary evidence, as it retains more trajectories than \textit{metric-driven} when one wants to achieve a dense representation of the set.

Trajectory sets are used for the downstream task of motion prediction of actors in the environment, where the probability of finding more trajectories outweighs the minimal performance gains from the displacement metrics. In this case, a set that maximizes plausibility is desired if the nominal gains from the Euclidean metrics are minimal, since we can guarantee that there exist more trajectories in the proximity which can represent the ground truth.

Although the \textit{metric-driven} method was able to provide guarantees of probabilistic completeness to maximize coverage, it did not capture the true distribution, providing insufficient exploration. On the other hand, sampling from a uniform distribution; \textit{RIDS} was able to retain more trajectories and outperform the \textit{metric-driven} method. The property of an efficient estimator is that it has the smallest possible variance, indicating that it is closest to the true value of the parameter being estimated. This reasoning that \textit{RIDS} being an efficient estimator and functions effectively because in certain optimization situations, sampling from the underlying distribution can be more efficient than performing sampling based on some metrics.

Our experiments have shown that the scene-specific set generation method allows us to improve diversity, and more specifically, \textit{RIDS}; a variant of random sampling works relatively well when considering the environment, resulting in an improved DAC score. 
The survival rate for the trajectories for sets constructed using \textit{Ran-Samp} and \textit{RIDS} outperforms \textit{metric-driven} methods, given that the minimal gain in displacement errors from \textit{metric-driven} overshadows the presence of more plausible trajectories. 
A visual distinction is notable for the non-intersection set in  \cref{fig:intersection_cluster}, which confirms our prior, as most of the trajectories do not represent significant yaw differences (going straight). The DAC score for the non-intersection with \textit{Ran-Samp} depicts that cluster-based trajectory sampling works better than expected, since it can retain 86\% of the trajectories for $T'_{5k}$. This high score indicates that the set can handle the complex geometry of the real-world and is perfectly suited for the intended application. As we said at the start, this showcases that Scene-Specific Set generation can produce sets which can be tuned to represent the ground truth closer to real-world scenes than using a single set to represent diverse situations.
Looking at \textit{Ran-Samp} set created with a threshold rate of 0.2 from \cref{table:Comparison_between_different_sets}; almost one could say that this imbalance between the set size is the reason for the improved DAC value. To counteract this argument, we performed ablation studies as can be seen from \cref{table:ran_samp@0.5} and \cref{table:ran_samp@0.2} to confirm that these improved DAC values remain improved when performing the RIDS sampling process for different threshold values.

% Begin Table ++++++++++++++++++++++++++++++++++
\begin{table*}[ht]
\centering
\caption{Ran-Samp @ threshold of 0.5}
\begin{tabularx}{1.75\columnwidth}{X|X|X|X|X|X|X}
\toprule
\textbf{Set size} & \textbf{After Threshold} & \textbf{Thresholded Set's LB. minADE} $(\downarrow)$ & \textbf{LB. minADE w/ $T_{full}$} $(\downarrow)$ & \textbf{DAC} $(\uparrow)$ & \textbf{RIDS (LB. minADE)} $(\downarrow)$  & \textbf{RIDS (DAC)} $(\uparrow)$\\ 
\midrule
1k   & 713  & 0.807 & 0.780   & 0.645 & 0.747   & 0.641 \\ 
1.5k & 993  & 0.770 & 0.710   & 0.623 & 0.694   & 0.619 \\ 
2k   & 1268 & 0.704 & 0.661   & 0.608 & 0.615  & 0.617 \\ 
2.5k & 1496 & 0.659 & 0.617   & 0.616 & 0.585  & 0.600 \\ 
3k   & 1684 & 0.647 & 0.598  & 0.608 & 0.564 & 0.591 \\ 
4k   & 2204 & 0.622 & 0.559  & 0.590 & 0.526  & 0.576 \\ 
5k   & 2645 & 0.597 & 0.532  & 0.582 & 0.497  & 0.562 \\ 
\bottomrule
\end{tabularx}
\label{table:ran_samp@0.5}
\end{table*}
% End Table ++++++++++++++++++++++++++++++++++

% Begin Table ++++++++++++++++++++++++++++++++++
\begin{table*}[t]
\centering
\caption{Ran-Samp @ threshold of 0.2}
\begin{tabularx}{1.75\columnwidth}{X|X|X|X|X|X|X}
\toprule
\textbf{Set size} & \textbf{After Threshold} & \textbf{Thresholded Set's LB. minADE} $(\downarrow)$ & \textbf{LB. minADE w/ $T_{full}$} $(\downarrow)$  & \textbf{DAC} $(\uparrow)$ & \textbf{RIDS (LB. minADE)} $(\downarrow)$ & \textbf{RIDS (DAC)} $(\uparrow)$\\ 
\midrule
1k   & 854  & 0.782 & 0.780   & 0.673 & 0.752   & 0.676 \\ 
1.5k & 1246 & 0.737 & 0.710   & 0.662 & 0.712   & 0.662 \\ 
2k   & 1642 & 0.666 & 0.661  & 0.664 & 0.635 & 0.680 \\ 
2.5k & 2067 & 0.621 & 0.617  & 0.678 & 0.599  & 0.676 \\ 
3k   & 2450 & 0.603 & 0.598  & 0.682 & 0.572  & 0.671 \\ 
4k   & 3237 & 0.564 & 0.559  & 0.671 & 0.534  & 0.670 \\ 
5k   & 4023 & 0.538 & 0.532  & 0.675 & 0.503  & 0.663 \\
\bottomrule
\end{tabularx}
\label{table:ran_samp@0.2}
\end{table*}
% End Table ++++++++++++++++++++++++++++++++++

%% file: sections/conclusion.tex
\section{Conclusion}
In this work, we presented our approach to create scene specific set generation which retains more trajectories akin to the environment which aims to maximize for the plausibility of trajectories for a diverse amount of environment. Along with that, we also showcased that RIDS as a sampling process is effective in condensing the full representation space to a more manageable set  while providing coverage and plausibility to the corresponding set.